\documentclass[sigconf]{acmart}

\usepackage{microtype}

\usepackage{graphicx}
\graphicspath{{tex_figs/}}

\usepackage{booktabs} 
\usepackage{hyperref}
\usepackage{algorithm}
\usepackage{algorithmic}
\usepackage{caption}
\usepackage{subcaption}
\usepackage{amsfonts}
\usepackage{makecell}
\usepackage{bm}

\newcommand{\R}{\mathbb{R}}
\newcommand{\E}{\mathbb{E}}
\newcommand{\I}{\mathbb{I}}
\newcommand{\cN}{{\mathcal N}}

\usepackage{amsthm}
\newtheorem{remark}{Remark}
\newtheorem{theorem}{Theorem}

\copyrightyear{2021}
\acmYear{2021}
\setcopyright{acmlicensed}
\acmConference[KDD '21] {Proceedings of the 27th ACM SIGKDD Conference on Knowledge Discovery and Data Mining}{August 14--18, 2021}{Virtual Event, Singapore.}
\acmBooktitle{Proceedings of the 27th ACM SIGKDD Conference on Knowledge Discovery and Data Mining (KDD '21), August 14--18, 2021, Virtual Event, Singapore}
\acmPrice{15.00}
\acmISBN{978-1-4503-8332-5/21/08}
\acmDOI{10.1145/3447548.3467080}

\begin{document}
\fancyhead{}

\title{Training Recommender Systems at Scale: Communication-Efficient Model and Data Parallelism}

\author{Vipul Gupta}
\email{vipul_gupta@berkeley.edu}
\affiliation{%
  \institution{University of California, Berkeley}
  \city{Berkeley}
  \state{CA}
  \country{USA}
}

\author{Dhruv Choudhary, Peter Tang, Xiaohan Wei, Xing Wang, Yuzhen Huang, Arun Kejariwal}
\affiliation{
\institution{Facebook Inc.}
\city{Menlo Park}
  \state{CA}
  \country{USA}
}

\author{Kannan Ramchandran, Michael W. Mahoney}
\affiliation{%
  \institution{University of California, Berkeley}
  \city{Berkeley}
  \state{CA}
  \country{USA}
}

\renewcommand{\shortauthors}{Gupta et al.}

\begin{abstract}

In this paper, we consider hybrid parallelism---a paradigm that employs both Data Parallelism (DP) and Model Parallelism (MP)---to scale distributed training of large recommendation models.
We propose a compression framework called Dynamic Communication Thresholding (DCT) for communication-efficient hybrid training. DCT filters the entities to be communicated across the network through a simple hard-thresholding function, allowing only the most relevant information to pass through.
For communication efficient DP, DCT compresses the parameter gradients sent to the parameter server during model synchronization. The threshold is updated only once every few thousand iterations to reduce the computational overhead of compression.
For communication efficient MP, DCT incorporates a novel technique to compress the activations and gradients sent across the network during the forward and backward propagation, respectively.
This is done by identifying and updating only the most relevant neurons of the neural network for each training sample in the data. 
We evaluate DCT on publicly available natural language processing and recommender models and datasets, as well as recommendation systems used in production at Facebook. 
DCT reduces communication by at least $100\times$ and $20\times$ during DP and MP, respectively. 
The algorithm has been deployed in production, and it improves end-to-end training time for a state-of-the-art industrial recommender model by 37\%, without any loss in performance. 

\end{abstract}


\maketitle


\section{Introduction}

\begin{figure*}[t]
    \centering
    \includegraphics[scale=0.25]{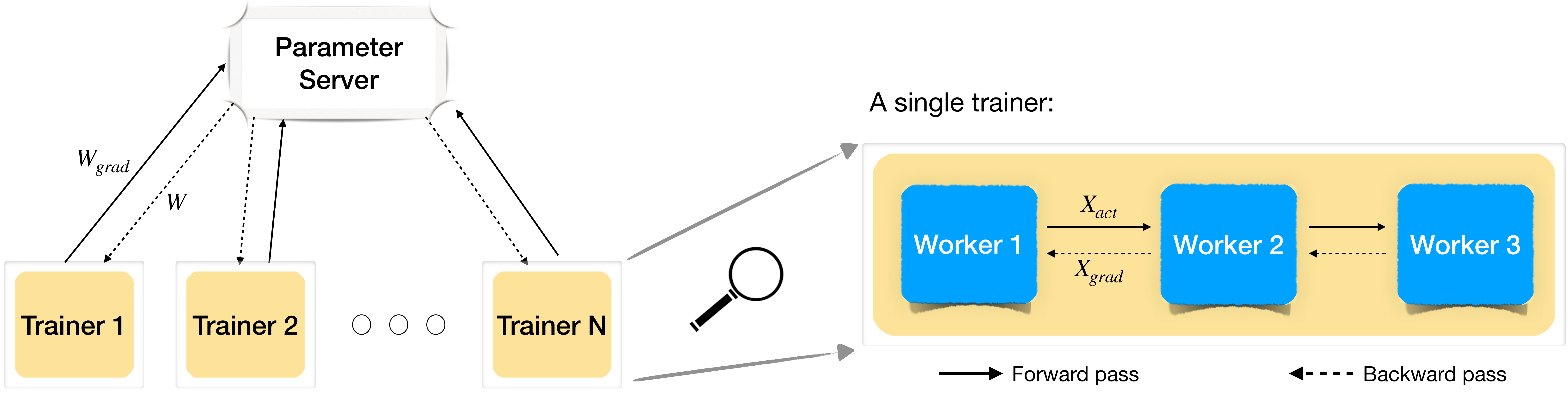}
    \caption{  Distributed DNN training with hybrid training which uses both DP (left) and MP (right) for greater parallelization gains.
    During DP, multiple trainers process several mini-batches of data in parallel. 
    During MP, one copy of the model is processed by one trainer which in turn is comprised of multiple workers.}
\label{fig:hybrid_illus}
\end{figure*} 

Data Parallelism (DP), in which each (of many) trainers stores a replica of the entire model, is a popular parallelization paradigm for the training of very large Deep Neural Networks (DNNs) 
\citep{dean2012large, DP_scaling}. 
At the beginning of each training iteration, each worker processes a subset of entire training data with a predetermined batch size, and then each worker synchronizes the model parameters at the end of the iteration. 
DP has experienced widespread deployment for state-of-the-art industrial applications, but it is now facing two major challenges. 
The first challenge is that large batch size is needed to exploit fully the ever-increasing compute power of training nodes. 
This turns out to be difficult. 
Both theoretical and empirical evidence suggests that going beyond a certain batch size for training DNNs results in loss in generalization performance (e.g., see \citep{keskar,LBnotEmp:2018shallue,openai,Hoffer17,LBempGolmant:2018,MM18_TR,ma2018_sgd_large_batch,staleness_async_sgd}). 
Despite active research on restoring generalization performance when the batch size is large \citep{priya2017,imagenet4mins:2018,lars,adabatch,smith2017,post-local-sgd,swap, yao2018hessian}, these methods either are specific to certain models and/or datasets, require extensive hyperparameter tuning, or can at best increase the maximum batch size by a small factor. 
The second challenge is replicating an entire DNN model on each worker, which is becoming an increasingly infeasible proposition. This is
due to increasing model complexity and parameters in domains such as, but not limited to, natural language processing and recommendation systems (e.g., see \citep{bert,outrageously_large_dnn,dense_conv_nets, dlrm}), coupled with the saturation of single machine memory and compute power due to trends such as the ending of Moore's law \citep{moore_law1, moore_law2}.

For these reasons, Model Parallelism (MP) has gained significant traction, both from the industry and the research community, as an alternative parallelization paradigm \citep{pipedream, gpipe,mp_multi_gpu,ElasticPipe,xpipe_async_mp,strads}. 
In its purest form, the entire network during MP is partitioned into a number of sub-networks equal to the number of workers. While this form can accommodate a larger network than DP, it fails to capitalize on the largest batch size that is allowable before generalization performance degrades.

Hybrid Parallelism (HP)---that employs both DP and MP---is a natural next step, an idea that was arguably first introduced in \cite{dean2012large}, and more recently exploited further for large-scale DNN training \citep{HT_google_2018,zaharia_hybrid_parallelism,hybrid_horizontal_vertical,hetpipe,optCNN, gholami2018integrated}. 
An illustration of hybrid training that uses MP to distribute the model across workers and DP to process multiple batches of training data at once is provided in Fig. \ref{fig:hybrid_illus}. 
Here, each partition of the network for MP is replicated in a group of workers, each processing the entire batch for that sub-network in question. Currently, hybrid training is employed in training a subset of large-scale recommendation models in production at Facebook.

The scaling of model size and batch size by HP has now progressed to the next bottleneck: communication bandwidth \cite{pipedream}.
This bottleneck exists in two crucial places. 
First, for MP, activation values and gradient information need to be communicated from one sub-network to the next during forward and backward propagation. 
Second, for DP, gradients of the same sub-network but for different sub-batches need to be communicated, regardless of the exact operations that follow.
This depends on the specific communication protocol (centralized versus decentralized reduction) or the algorithm (synchronous versus asynchronous updates). 
To compound the problem, increasing the batch size to fully exploit DP increases the communication of activations and gradients in MP, the sizes of which are directly proportional to the batch size. 
Additionally, in the asynchronous training, increasing batch size exacerbates the stale gradient problem due to an increase in the time interval between a worker receiving the model and sending the gradient \cite{revisiting_sync_sgd}. 
In short, the benefits of communication reduction are many.

\textbf{Dynamic Communication Thresholding.} 
We propose a Dynamic Communication Thresholding (DCT) framework for communication efficient training for HP. 
DCT incorporates two algorithms, DCT-DP and DCT-MP, to alleviate communication congestion for DP and MP, respectively. 
Our algorithms filter the entities to be communicated through a simple hard-thresholding function, eliminating the need to pass many of them over the communication fabric. 
We propose practical methods to compute the thresholds to reduce the computational overhead of compression. 
Our thresholding technique is versatile, as it applies to different communication primitives in DP for the gradients, to different pipelining methods in MP (e.g., GPipe \citep{gpipe}, PipeDream \citep{pipedream}), and to different applications such as recommendation systems and natural language processing models. 
While thresholding communication introduces errors, we apply (previously known) error compensation technique as well as a model consistency adjustment method (we developed) to mitigate the effect of the error in compression. 
Consequently, despite significant communication thresholding, model accuracy does not degrade, and in fact it often improves. 

We apply DCT to large-scale state-of-the-art recommendation models in production with real-world datasets as well as publicly available models and datasets. 
We observe that the communication costs are reduced by factors of up to $20\times$ for MP and $100\times$ for DP. Further, end-to-end training time for large-scale training is cut by as much as 37\% for industry-scale models in production. 
Further, applying DCT reduces the network utilization from 94.2\% to 49.3\% and increases the overall CPU utilization from 48.7\% to 91.1\%, shifting the bottleneck of model training from communication to computation in such systems.

\textbf{Related Work.}
Due to the use of large clusters with powerful machines to train complex DNNs (e.g. BERT-Large \citep{bert} with 340M parameters), the distributed training workloads are becoming increasingly communication bound. For this reason, numerous compression schemes have been proposed in the past several years for the data parallel setting (see \citep{DP_survey} for a comprehensive survey). 
These compression schemes come in various forms, such as the following: 
(i) Quantization, where the number of bits per entry of the communicated vector is reduced (e.g., \citep{alistarh2017_qsgd, EF_signSGD_karimireddy, terngrad}); 
(ii) Sparsification, where only a few entries of the communicated vector are sent (e.g., \citep{stich_sparsified_sgd,alistarh2018_conv_sparse,aji2017sparse,deep_grad_comp_lin, sun2019sparse, wangni2018sparse}); 
(iii) Statistical techniques such as Randomized Sketching (e.g., \citep{sketchml,sketched-SGD}); and 
(iv) Low-rank approximation, which decomposes the vector into low-rank components before communication (e.g., \cite{lowrank_atomo,lowrank_powersgd,lowrank_gradzip,lowrank_gradiveq}). 

When it comes to performance on real-world systems, many of these existing schemes have one or more of the following shortcomings. 
(i) Focus is mostly on a theoretical analysis of schemes based on restricted assumptions, such as convexity and synchronous SGD.  
(ii) The empirical evaluation ignores the cost of compression and decompression which, in many cases, deprives them of any savings due to communication. 
(iii) Comparison of convergence with respect to baseline is reported, while the number of epochs (or iterations) and the actual training time is ignored. 
For instance, in Fig. 1 in \cite{DP_survey}, the authors compare the compression scheme in \citep{sketchml} with a baseline without compression.
They observe that, although the convergence with respect to the number of epochs is unaffected due to compression, it takes almost twice the time for training to converge, rendering the scheme worse than no compression.
We also observed in our experiments that for sparsification using top-K sparsity \cite{stich_sparsified_sgd,alistarh2018_conv_sparse}, the overhead of copying and sorting the large vectors ends up taking more time than the gains obtained due to communication reduction. 
(See Fig. \ref{fig:DCT-DP-Prod} in Sec. \ref{sec:exps_prod} for details.) 

In this paper, {\bf we propose practical schemes for communication reduction during DP, and we show performance improvements in terms of the end-to-end DNN training times}, with performance similar to, or in some cases better than, the baseline algorithms as implemented in industry.
For the MP case, existing works target the scheduling of communication of entities across the network to improve the efficiency of training DNNs \citep{lee2014_MP_sch, DES_MP}. However,
to the best of our knowledge, {\bf this is the first work that targets communication reduction for MP by compressing the entities (i.e., activations and gradients) that are sent across the network}. 
As such, it can be applied on top of existing training efficiency schemes, such as communication scheduling \citep{lee2014_MP_sch, DES_MP} and Pipelining \citep{pipedream, gpipe, pipemare, xpipe_async_mp} for MP. As illustrated in Fig. \ref{fig:hybrid_illus} (right), communication is a major bottleneck for MP-based training since the activations are communicated from (say) worker 1 to worker 2 during the forward pass and the gradients are then communicated from worker 2 to worker 1 during the backward pass (similar communication happens between workers 2 and 3). 
However, we further observed that naively applying compression schemes, such as sparsification, quantization and sketching, to the activations and gradients either do not achieve high enough compression rates to be practical, or the degradation in model performance is beyond an acceptable level.
(See Appendix \ref{app:sketch} for details on such negative results.)

In the next section, we describe our algorithms for communication efficiency during parallelization, for both the MP and DP primitives of the DNN training.
In particular,
we discuss DCT-DP (in Section \ref{sec:dct_dp}) and explain our gradient reduction technique for DP that requires minimal computational overhead for compression; and then we discuss DCT-MP (in Section \ref{sec:dct_mp}), a flexible thresholding framework with theoretical support for our design.
Then, Section \ref{sec:exps} reports our findings from a diverse set of experiments and demonstrates the advantages of using DCT-DP and DCT-MP for training large-scale models for both publicly available and production models.


\section{Communication-Efficient Training with Hybrid Parallelism}

We start, in Section \ref{sec:dct_dp}, by proposing a Dynamic Communication Thresholding (DCT) technique for DP (DCT-DP). 
DCT-DP is inspired by existing theoretical works such as \citep{stich_sparsified_sgd} and \cite{alistarh2018_conv_sparse}. 
It sparsifies the gradient in each iteration before sending it over the wire, and it intelligently chooses the threshold for sparsification to reduce the computational overhead introduced due to compression and decompression. 
Then, in Section \ref{sec:dct_mp}, we propose DCT-MP, a novel thresholding scheme for sparsification of activations and gradients during forward and backward passes, respectively, to reduce communication during MP.

\subsection{DCT-DP: Reducing communication for Data Parallelism}\label{sec:dct_dp} 
During DP, as illustrated in Fig. \ref{fig:hybrid_illus} (left), we compress the gradient, $W_{grad}$, from trainers to the parameter server to improve the communication bottleneck.
Our compression algorithm, DCT-DP, is inspired by previous works which focus on data-parallel training for alleviating communication bottlenecks, and in particular the works of \citep{stich_sparsified_sgd, alistarh2018_conv_sparse}, where error feedback is employed along with sparsification to correct the error in gradient direction due to compression. 
Such schemes find a top-K threshold by sorting the gradient vector, and they use the threshold to sparsify the gradients by keeping only the top-K entries.
However, they focus on proving theoretical convergence guarantees, and they do not show improvements in end-to-end times for training neural networks. 

In our experiments, we observed that the overhead of allocating memory to copy the gradient (with its size easily scaling into the millions) and sorting the resultant copy to find the top-K threshold in each iteration is sufficiently expensive that it deprives any improvements in end-to-end training time in real-world systems (see Sec. \ref{sec:exps_prod} for details). 
Hence, such gradient compression schemes, in their most basic form, cannot be employed directly to obtain promised gains in training efficiency. However, we take advantage of the following observation to reduce the overhead introduced due to compression.

\begin{figure}[t]
 \centering
 \includegraphics[width=0.45\textwidth]{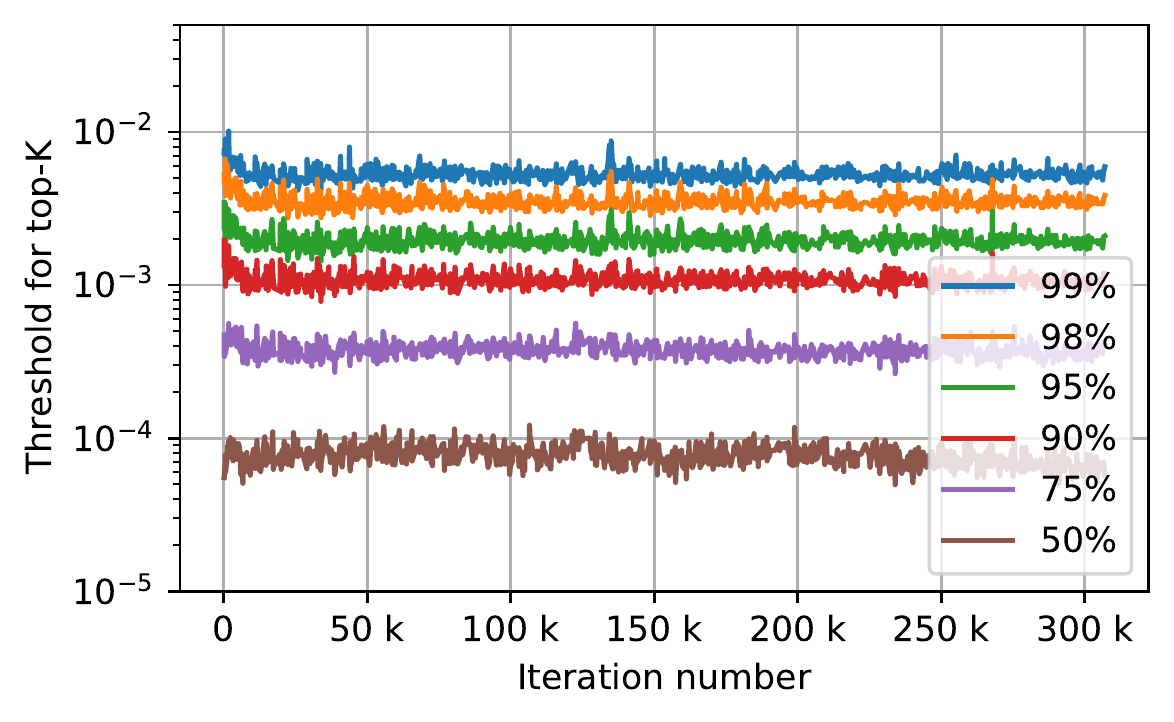}
 \caption{  Top-K threshold for various levels of sparsity during the gradient compression for DCT-DP. We see that the top-K thresholds, for different sparsity levels, do not deviate much from the mean. Thus, updating the threshold only every $L (> 1)$ iterations can help reduce the overhead of sorting to find the top-K threshold.}
 \label{fig:thres_for_w_grad}
\end{figure}

In Fig. \ref{fig:thres_for_w_grad}, we plot the top-K thresholds for various levels of sparsity for the Deep Learning Recommendation Model (DLRM) \citep{dlrm} with the Criteo Ad Kaggle Dataset for one of the Fully Connected (FC) layers (see Sec. \ref{sec:exps_dlrm} for details on the training process). We see that the threshold value increases as the sparsity increases, which is expected. More importantly, we note that given a sparsity factor, the threshold value does not vary much across iterations. For example, for $95\%$ sparsity, the threshold deviates by at most $26\%$ around its running mean. Thus, even for reasonably large compression factors, updating the threshold every iteration is excessive.

\begin{figure}[t]
 \centering
 \includegraphics[width=0.45\textwidth]{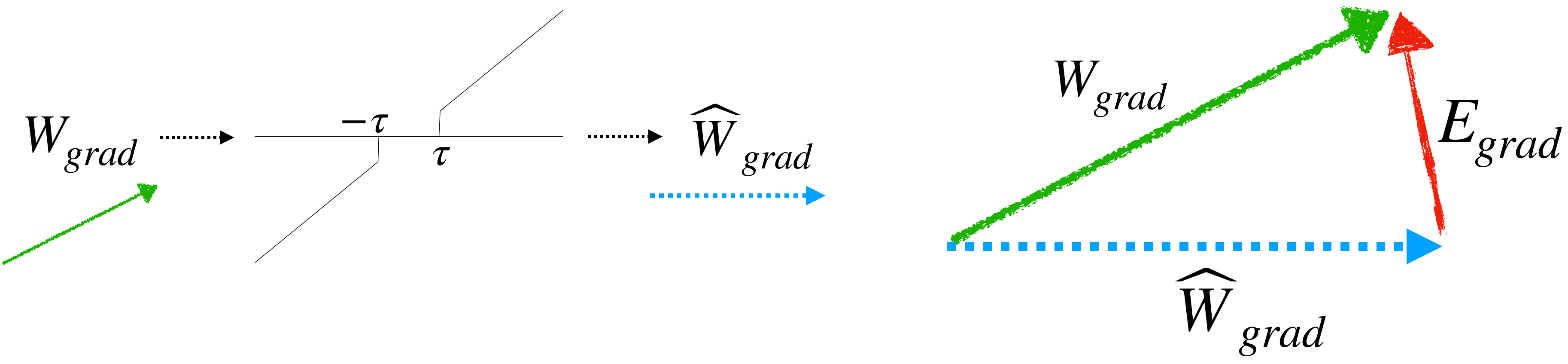}
 \caption{  A illustration of DCT-DP. 
 First, $W_{grad} \in \R^N$ (which already incorporates error from the previous iteration) is compressed using a threshold $\tau$ to obtain the sparse vector $\widehat W_{grad}$. 
 Then, the error is calculated as $E_{grad} = W_{grad} - \widehat W_{grad}$ to be used in the next iteration to correct the error in gradient direction.}
 \label{fig:wta_dp}
\end{figure}

Inspired by this observation, we update the threshold only once every $L$ iterations (where $L$ is generally in thousands) while compressing the gradient of the parameters, $W_{grad}$, for each DNN layer. We refer to $L$ as the threshold life-span. As we observe in our experiments (see Sec. \ref{sec:exps_prod}), we can compress the gradients by as much as $99\%$ sparsity with $L=1000$ for each layer using top-K sparsification and error correction without any loss in performance. Our algorithm is illustrated in Fig. \ref{fig:wta_dp} and detailed steps are provided in Algorithm \ref{alg:wta_dp}. Throughout this paper, the function $\I(\cdot)$ denotes the indicator function, and the symbols $\lfloor\cdot\rfloor$ and $\odot$ denote the integer floor and element-wise product of two matrices, respectively.

Note that each trainer consists of multiple workers, and each worker compresses the gradients layer-wise using sparsification before communication (see Fig. \ref{fig:hybrid_illus} for an illustration, where each trainer consists of 3 workers). This is unlike existing works (e.g. \citet{stich_sparsified_sgd, sketched-SGD}) where the gradient vectors of all the model parameters are combined and compressed together. However, the theoretical guarantees on the convergence of the algorithm still holds and can be trivially extended to our case. This is because, for any threshold $\tau > 0$, the compressed gradient satisfies the contraction property (Definition 2.1 in \citet{stich_sparsified_sgd}). Hence, DCT-DP satisfies the same rate of convergence as Stochastic Gradient Descent (SGD) without compression (see Theorem 2.4 in \citet{stich_sparsified_sgd}). 

\begin{algorithm}[t]
   \caption{DCT-DP: Communication-Efficient Data Parallelism}
\begin{algorithmic}[1]
   \STATE {\bfseries Input:} Sparsity factor $\eta$ ($0 < \eta \leq 1$), Threshold life-span $L$, Iteration number $k$, Gradient of the DNN layer $W_{grad} \in \R^N$, Error $E_{grad} \in \R^N$, and Threshold $\tau$ (from iteration $k-1$)
   \STATE {\bf Error Feedback}: $W_{grad} = W_{grad} + E_{grad}$ 
   \IF{$L$ divides $k$}
   \STATE $[w_1, w_2, \cdots, w_N]$ = $Sort(|W_{grad}|)$ 
   \STATE Assign $\tau = w_{\lfloor N\times \eta \rfloor}$ 
   \ELSE 
   \STATE Use $\tau$ from iteration $k-1$
   \ENDIF
   \STATE Compute mask $M = \I(|W_{grad}| \geq \tau)$
   \STATE Compute compressed gradient $\widehat W_{grad} = W_{grad} \odot M$
   \STATE Compute error $E_{grad} = W_{grad} - \widehat W_{grad}$
   \STATE Send $\widehat W_{grad}$ to the parameter server which updates the model
\end{algorithmic}
\label{alg:wta_dp}
\end{algorithm}

\subsection{DCT-MP: Reducing communication for Model Parallelism}\label{sec:dct_mp}

\begin{figure}[t]
 \centering
 \includegraphics[width=0.45\textwidth]{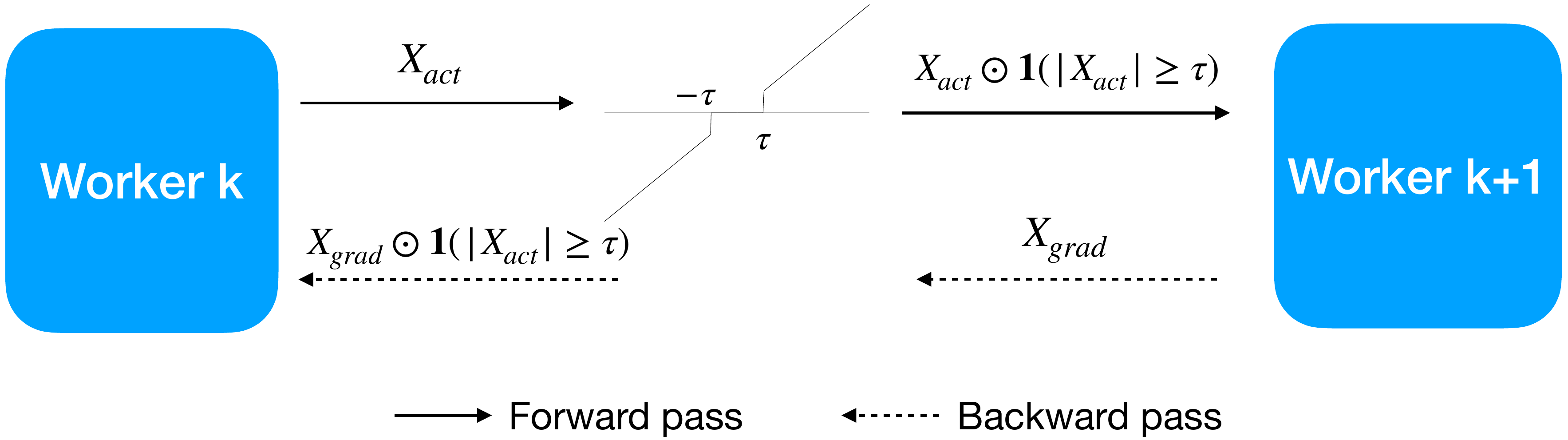}
 \caption{  A illustration of DCT-MP. During the forward pass, we sparsify and compress the activations, say $X_{act}$, corresponding to one data sample, using the mask, $\I(|X_{act}| \geq \tau)$, is generated based on the threshold $\tau$. During the backward pass, the same mask is used to compress the gradients and selectively train neurons.}
 \label{fig:wta_mp}
\end{figure}

Training of large-scale DNNs is often regarded with pessimism due to its associated training latency (multiple days/weeks). 
However, training such large-scale models can be a ``blessing in disguise'' from a communication-efficiency point of view. 
For such models, with billions of parameters in each layer, only a few of the neurons are activated during the forward pass, potentially allowing us to compress these activations by a factor of $20\times$ or more with no loss in model performance. 
This idea of training only a subset of neurons every iteration based their activation values stems from several existing observations \cite{makhzani_k_sparse2013,makhzani2015_wta_autoencoder,slide}.
In fact, in works such as dropout \cite{dropout} and adaptive dropout \cite{adaptive_dropout}, the authors have shown that selective sparsification can improve the generalization performance due to implicit regularization \citep{Mah12}. 
With such a scheme, we also observe gains in generalization performance on top of communication efficiency (see experiments in Section \ref{sec:exps}).

Motivated by this, we propose a sparsification scheme where the neurons compete with each other in every iteration during DNN training, and the ones with the largest (absolute) value of activations are selected. 
Thus, for a given training sample, DCT-MP selects only a few neurons (say $\sim$5\%) during the forward pass that are generally sufficient to represent the entire information for that training sample. We next describe DCT-MP in more detail.

{\bf Algorithm.} 
Let the mini-batch size be $B$ and the number of output features before the model split be $d$. Thus, the activation and gradient matrices ($X_{act}$ and $X_{grad}$, respectively) lie in $\R^{B\times d}$. Based on the idea that each example activates only a subset of neurons, we select a fraction, say $\eta$, of largest entries according to their absolute value in each row. Thus, for the $i$-th row of $X_{act}$, say $X_{act,i}$, we select a threshold $\tau_i$ which is greater than $d\times\eta$ values in $X_{act,i}$, and the mask is thus calculated for the $i$-th data sample as $\I(X_{act,i} \geq \tau_i)$. The same mask is then used to compress the entities $X_{act,i}$ and $X_{grad,i}$ during forward and backward passes, respectively, for all $i\in \{1,2, \cdots, B\}$. Thus, the training for each mini-batch happens only on the relevant neurons corresponding to each sample in the training data.
In Fig. \ref{fig:wta_mp}, we illustrate the compression using DCT-MP when the mini-batch size is one. Detailed steps for a general mini-batch size $B$ are provided in Algorithm \ref{alg:wta_mp}. 

\begin{algorithm}[tb]
   \caption{DCT-MP: Communication-Efficient Model Parallelism}
   \label{alg:wta_mp}
\begin{algorithmic}[1]

   \STATE {\bfseries Input:} Sparsity factor $\eta$ ($0 < \eta \leq 1$),

   {\underline{Forward Pass}}:
   \STATE {\bfseries Input:} Activation matrix $X_{act} = [X_{act,i}]_{i=1}^{B} \in \R^{B\times d}$
   \STATE Define the mask, $M = [~]$
   \FOR{$i=1$ {\bfseries to} $B$}
   \STATE $[x_1, x_2, \cdots, x_d]$ = $Sort(|X_{act,i}|)$ 
   \STATE Define $\tau_i = x_{\lfloor d\times \eta \rfloor}$ 
   \STATE $m_i = \I(|X_{act,i}| \geq \tau_i)$
   \STATE $M = [M; ~m_i]$
   \ENDFOR
   \STATE Compute the sparse matrix $X_{act}\odot M$
   \STATE Send $X_{act}\odot M$ and the mask $M$ across the network

   {\underline{Backward Pass}}:
   \STATE {\bfseries Input:} Gradient matrix $X_{grad} \in \R^{B\times d}$
   \STATE Compute the sparse matrix $X_{grad}\odot M$
   \STATE Send $X_{grad}\odot M$ across the network
\end{algorithmic}
\end{algorithm}

\begin{figure}[t]
 \centering
 \includegraphics[width=0.44\textwidth]{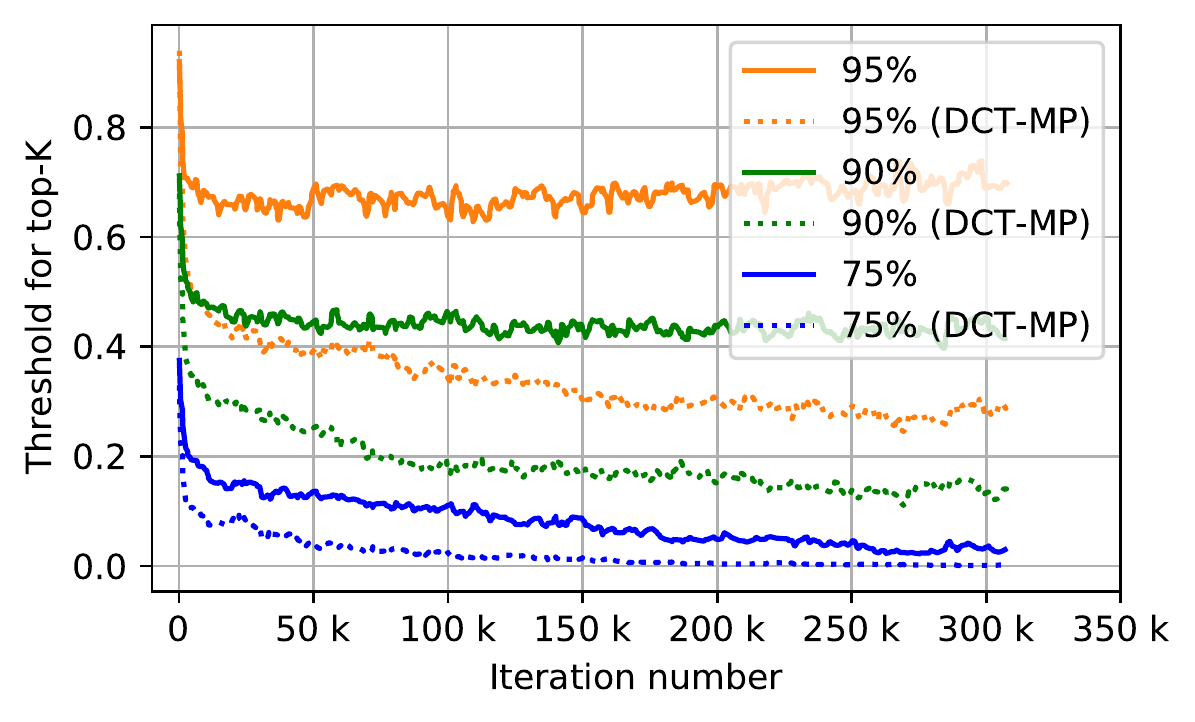}
 \caption{  Top-K threshold for various levels of sparsity for the cases when compression using DCT-MP is applied and when it is not applied. The top-K thresholds decrease significantly when DCT-MP is applied. Thus, DCT-MP induces sparsity in neuron activations.  This is possibly the reason for its improved generalization performance.}
 \label{fig:thres_values_for_MP}
\end{figure}

{\bf DCT-MP Promotes Sparsity in Model Activations.} 
In Fig. \ref{fig:thres_values_for_MP}, 
we plot the mean, $\frac{1}{B}\sum_{i=1}^B \tau_i$, of threshold vector $\tau = [\tau_1,\tau_2, \cdots, \tau_B]$ with respect to the number of iterations for the DLRM model with the Criteo Ad Kaggle Dataset. The threshold is calculated for activations after one of the fully connected layers (see Sec. \ref{sec:exps_dlrm} for details on the experimental setup). The mean of the threshold is calculated for different sparsity levels ($75\%, 90\%$ and $95\%$) for the two cases when sparsification using DCT-MP is applied (dotted lines) and when it is not applied (solid lines). Thus, the solid lines correspond to a single training run where we are simply measuring the mean of top-K threshold values without actually sparsifying the activations sent across the wire. The dotted lines with different sparsification levels correspond to different training runs where the stated sparsification is actually applied to the activations (and gradients) that are sent across the wire.  

We observe that, as the training progresses, the top-K thresholds decrease significantly faster for the case when DCT-MP is applied. A decrease in the top-K threshold corresponds to the activations getting sparser (maybe approximately) as the training progresses. Thus, DCT-MP induces sparsity in activations while training, which is exploited for communication efficiency. An important advantage of such sparsity-inducing regularization is the improved generalization performance of the model, as shown in our experiments in Sec. \ref{sec:exps}. Our conjectured explanation for why sparsity helps in improving the generalization error is based on the performance of existing popular schemes.
This includes dropout (see Fig. 8, \citet{dropout}) and Rectified Linear Units (ReLU) (see Fig. 3, \citet{relu}), which themselves introduce sparsity in model activations, as well as implementations of implicit sparsity based methods in scalable algorithms for graph analysis \citep{SRFM16_VLDB,FGM17_IEEE}.


{\bf Analysis of DCT-MP.} 
To provide further insights into DCT-MP, we prove that the stochastic gradient obtained with Algorithm \ref{alg:wta_mp} is equal, in expectation, to the stochastic gradient obtained in a network without any communication thresholding. More details of this unbiased estimation, including a formal statement of the theorem and its proof, are provided in Appendix \ref{app:analysis_dct_mp}. 

{\bf Comparision with Dropout.} Dropout and DCT-MP are similar in essence as they both selectively train neurons. However, the two schemes are different: both in the goals they try to achieve, and in the mechanisms they use. Furthermore, they can be used complementarily. Here are the main differences between the two schemes. First, Dropout drops neurons randomly, while DCT-MP keeps only the most relevant neurons for each training sample. Second, for Dropout, going beyond 50\% sparsity results in accuracy loss, but DCT-MP achieves up to 95\% sparsification. Third, Dropout is applied to every parameter layer, but DCT-MP is applied only to the layers before the model split.

\section{Empirical Results}
\label{sec:exps}

In this section, we investigate DCT-MP and DCT-DP for three different experimental setups. In subsections \ref{sec:exps_dlrm} and \ref{sec:exps_nlp}, we evaluate the performance of DCT-MP on the Deep Learning Recommendation Model (DLRM) and a Natural Language Processing (NLP) model, respectively, for different levels of compression and different number of MP workers. The models and datasets are publicly available. We show that high compression factors can be obtained (up to $\sim$$95\%$) with DCT-MP along with small improvements in model performance. 

We further evaluate DCT-DP on the DLRM model in subsection \ref{sec:exps_dlrm} and see no loss in performance with up to $98\%$ sparsity.
Finally, we evaluate the performance of DCT-DP and DCT-MP on large-scale recommendation models that are trained with hybrid parallelism in production systems. 
We show that the deployed algorithm reduces the training time by $37\%$ for such production-scale models without any performance loss.

Further, in all our experiments, we tried to show at least one negative result that would provide insights into the scalability of DCT. For instance, as the number of workers for MP (i.e., model splits) increases, the compression factor with DCT-MP decreases (e.g., Tables \ref{table:MP_DLRM}, \ref{table:DP_MP_DLRM}, \ref{table:DP_MP_NLP} and \ref{table:prod_mp}).

\subsection{Experiments on the DLRM Model}
\label{sec:exps_dlrm}

{\bf Experimental Setup.} 
For these experiments, we use the DLRM model from \citep{dlrm}. In this model, the dense features are first processed by a Multilayer Perceptron (MLP) with four layers, where each layer contains a Fully Connected (FC) layer followed by a Rectified Linear Unit (ReLU). Then, there is a feature interaction between the processed dense and sparse features, which goes through a second MLP with four layers (the last layer has Sigmoid instead of ReLU as the non-linearity) to produce the final output. 
In our experiments, the embedding dimension for sparse features was kept at 16, and the output dimensions of the four FC layers in the first MLP are 512, 256, 64 and 16, respectively. Similarly, for the second MLP, the output dimensions for the fours FC layers are 512, 256, 128 and 1, respectively.%
\footnote{See the Criteo Kaggle benchmark for further details on the training process: https://github.com/facebookresearch/dlrm} 
Training and testing sets comprise of 6 days and one day, respectively, of the Criteo Ad Kaggle dataset.%
\footnote{https://labs.criteo.com/2014/02/kaggle-display-advertising-challenge-dataset/} 

\begin{figure*}[t]
 \centering
 \includegraphics[width=0.85\textwidth]{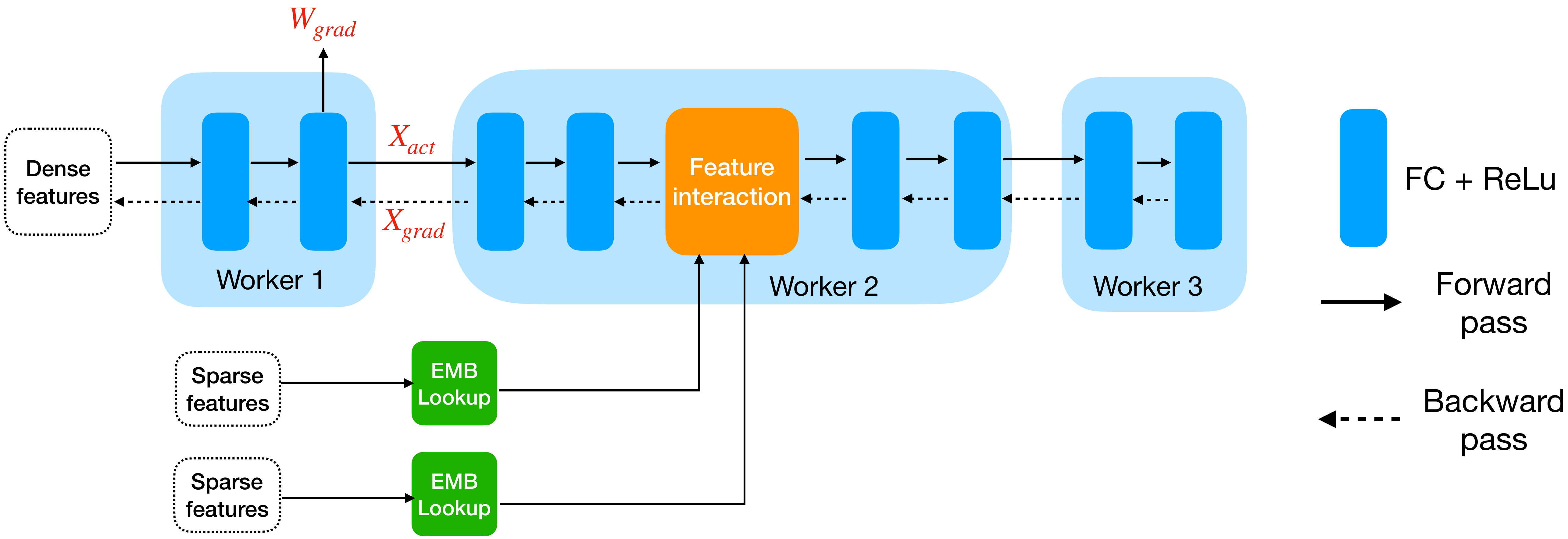}
 \caption{  A illustration of model parallelism with DLRM. The entities that are sent across the network are shown in red. $X_{act}$ and $X_{grad}$ are communicated during MP, and $W_{grad}$ is communicated during DP. The shaded area in blue represents a sample model partitioning for MP. In this case, three workers are working on one copy of the model during MP and comprise a trainer.}
 \label{fig:dlrm_illus}
\end{figure*}

Fig. \ref{fig:dlrm_illus} provides an illustration of MP with the DLRM model. The shaded area in blue shows a sample partition for MP. 
In our simulations, we consider up to two splittings of the DLRM model. The first split is after two layers in the first MLP, and the second split is after two layers in the second MLP. Our goal is to reduce communication across different workers (both during the forward and backward passes). This is a typical setup in MP Training where workers 1, 2, and 3 can be the different pipes of a single trainer (e.g., see \citet{gpipe}).
For all our experiments, the data shuffle remains constant across different training runs.

In Fig. \ref{fig:dlrm_illus}, we mark the three entities that are sent across the network which we compress to alleviate communication costs in distributed DNN training. $X_{act}$ and $X_{grad}$ are the activation and gradient matrices sent across the network during the forward pass and backward passes, respectively. The third entity that can be compressed is the parameter gradient (shown as $W_{grad}$) that is sent from Workers 1, 2, and 3 to the parameter server.
This keeps a central copy of weights and updates it regularly through the gradients received from different workers.

{
\begin{table}[ht]
\caption{  DCT-MP on the DLRM model: Train and Test Loss and Accuracy for multiple sparsity ratios (denoted by $\eta$) and different settings for MP.}
\label{table:MP_DLRM}
\begin{center}
\begin{small}
\begin{sc}
\begin{tabular}{lccccr}
\toprule
\makecell{$\eta$}  &
\makecell{MP\\Workers} &
\multicolumn{2}{c}{
\makecell{Train\\Loss \phantom{iii} Acc (\%)}} &
\multicolumn{2}{c}{
\makecell{Test\\Loss \phantom{iii} Acc (\%)}} \\
\midrule
0\% & -- & 0.4477 & 79.23 & 0.4538 &  78.78\\
\cmidrule{1-6}
75\%    & 2  & 0.4473 & 79.29 & 0.4532 &  78.81\\
90\%    & 2  & 0.4472 & 79.28 & 0.4530 &  78.81\\
{\bf 95\%}    & 2  & 0.4473 & 79.24 & \bf 0.4534 & 78.80\\
98\%    & 2  & 0.4505 & 79.07 & 0.4562 &  78.61\\
\cmidrule{1-6}
75\%    & 3  & 0.4482 & 79.19 & 0.4536 &  78.79\\
\bf 90\%    & 3  & 0.4479 & 79.24 & \bf 0.4537 & 78.78\\
95\%    & 3  & 0.4495 & 79.18 & 0.4546 &  78.72\\
\bottomrule
\end{tabular}
\end{sc}
\end{small}
\end{center}
\end{table}
}

In Table \ref{table:MP_DLRM}, we show the cross-entropy loss \cite{DL_book} and accuracy with the DLRM model on the training and testing data samples. A sparsity factor ($\eta$) of $0\%$ denotes the baseline with no compression. 
We consider two settings for MP: one split (that is, 2 MP workers); and two splits (or three workers for MP).

{\bf MP with two workers (one split).} 
In rows 2-5 in Table \ref{table:MP_DLRM}, we consider one split in the model (or MP with two workers) in the first MLP after two layers. We see that even with $95\%$ sparsity (that is, $20\times$ compression) on $X_{act}$ (and $X_{grad}$) sent across the network, we are able to perform better than baseline (with no compression), both in terms of train and test loss (highlighted in bold cases). However, we see a tangible loss in performance when the sparsity is further increased to $98\%$. 


{\bf MP with three workers (two splits).} 
In rows 6-8 in Table~\ref{table:MP_DLRM}, we consider MP with 3 workers, where the two model splits are in the first and second MLP, as shown in Fig. \ref{fig:dlrm_illus}. Note that, in the case of two splits, compressing the entities that are sent across the network by up to $90\%$ does not affect the test accuracy, and it is still better than the baseline with no compression.  However, increasing the sparsity factor to $95\%$ is too ambitious for the two split case, and it increases the test loss by $0.18\%$. Further increasing the number of splits results in a greater performance loss, and the performance is worse than baseline for even $75\%$ sparsity.

\begin{remark}
We emphasize that for all the experiments in this paper, the location of splits for MP were not tuned as hyperparameters. Instead, we inserted splits after randomly chosen FC layers, or after the ReLU following the FC layer if it exists. The advantage of inserting a split after ReLU layers is that the activation matrix is $50\%$ sparse on average, resulting in higher compression rates for DCT-MP. 
\end{remark}

\begin{table}[ht]
\caption{  DCT-DP on the DLRM model: Train and Test Loss and Accuracy for various levels of sparsity.}
\label{table:DP_DLRM}
\begin{center}
\begin{small}
\begin{sc}
\begin{tabular}{lcccr}
\toprule
\makecell{Sparsity \\ Factor} & 
\multicolumn{2}{c}{
\makecell{Train\\Loss \phantom{iii} Acc (\%)}} &
\multicolumn{2}{c}{
\makecell{Test\\Loss \phantom{iii} Acc (\%)}} \\
\midrule
Baseline  & 0.4477 & 79.23 & 0.4538 &  78.78 \\
\cmidrule{1-5}
75\%      & 0.4478 & 79.23 & 0.4534 &  78.81\\
90\%      & 0.4478 & 79.22 & 0.4536 &  78.79\\
95\%      & 0.4479 & 79.25 & 0.4538 &  78.79\\
\bf 98\%      & 0.4478 & 79.23 & \bf 0.4537 & 78.80\\
99.5\%      & 0.4482 & 79.20 & 0.4547 &  78.75\\
\bottomrule
\end{tabular}
\end{sc}
\end{small}
\end{center}
\end{table}

{\bf DP with the DLRM Model.} 
In Table \ref{table:DP_DLRM}, we illustrate the performance of DCT-DP on DLRM by compressing the gradients of the parameters of all the 8 FC layers while they are sent across the wire to the parameter server. 
The parameter server then updates the model parameters using the compressed gradient. We use error feedback \cite{EF_signSGD_karimireddy} to compensate for the error in gradient compression by feeding it back to the gradients in the next iteration. In general, DCT-DP compression enjoy higher compression rates due to the use of error compensation schemes and the fact that error in one layer does not propagate to the other layers, unlike in the case of MP compression.  Compression up to $98\%$ sparsity does not show any loss in performance. However, further compressing to $99.5\%$ sparsity increases the test loss by $0.20\%$.

\begin{table}[ht]
\caption{  Compression using DCT-DP and DCT-MP on the DLRM model: Train and Test Loss and Accuracy with two MP splits (that is, three workers for MP).}
\label{table:DP_MP_DLRM}
\begin{center}
\begin{small}
\begin{sc}
\begin{tabular}{lcccr}
\toprule
\makecell{Sparsity \\ Factor} & 
\multicolumn{2}{c}{
\makecell{Train\\Loss \phantom{iii} Acc (\%)}} &
\multicolumn{2}{c}{
\makecell{Test\\Loss \phantom{iii} Acc (\%)}} \\
\midrule
Baseline  & 0.4477 & 79.23 & 0.4538 &  78.78 \\
\cmidrule{1-5}
75\%      & 0.4480 & 79.23 & 0.4535 &  78.81\\
\bf 90\%      & 0.4481 & 79.26 & \bf 0.4537 &  78.78\\
95\%      & 0.4492 & 79.19 & 0.4548 &  78.70\\
\bottomrule
\end{tabular}
\end{sc}
\end{small}
\end{center}
\end{table}

{\bf Communication-efficient Hybrid Training.} 
Next, we apply compression to $W_{grad}$ for the 8 FC layers (in the DP case) and to $X_{act}$ (and $X_{grad}$) for two splits (in the MP case) and present our results in Table \ref{table:DP_MP_DLRM}. We see that compression up to $90\%$ sparsity (both during DP and MP) does not affect the performance, but the test loss increases by $0.22\%$ when the sparsity factor is increased to $95\%$.

\subsection{Experiments on a Translation Model}
\label{sec:exps_nlp}

For our experiments with DCT-MP, we next consider the Transformer translation model as an application of NLP using DNNs. We train over the IWSLT'14 German to English dataset \citep{iwslt}.
The setup and hyperparameters were directly borrowed from the fairseq NLP Library \cite{fairseq}.  
The model used was borrowed from \cite{vaswani2017attention}, where both encoder and decoder have 6 layers, each of which uses a fully connected Feed-Forward Network (FFN) with input and output dimensionality of 512 and inner layer dimensionality of 1024.%
\footnote{For further details on the translation model, dataset preprocessing and the hyperparameters used, see https://github.com/pytorch/fairseq/tree/master/examples/translation} 
We report the training and testing losses and the BLEU scores after 50 epochs of training. 

Our results with DCT-MP on the translation model are described in Table \ref{table:DP_MP_NLP}. We consider three training scenarios: Two MP workers (with one split), Three MP workers (with two splits), and Five MP workers (with 4 splits). For the case with one split, we inserted the DCT-MP operator after the ReLu operator in the FFN of the fifth encoder layer. For the two splits case, we additionally inserted the DCT-MP operator after the ReLu operator in the FFN of the fifth encoder layer. We further added two splits after the ReLu operator in the third FFN in both the encoder and decoder layers for the four splits case. For each scenario, we show the best performing sparsity factor in bold.   

We emphasize that no hyperparameter tuning was performed in choosing the splits, and we observed in our experiments that using DCT-MP after an FC Layer or a ReLu layer improves the generalization performance, possibly due to (implicitly) added regularization (as illustrated in Fig. \ref{fig:thres_values_for_MP}). 
Note that we can add more MP splits for the NLP model compared to the DLRM model since the model is significantly deeper (and thus less susceptible to changes in outputs of a few layers) with larger FC layers (thus allowing for greater sparsity). 
This shows that DCT-MP is more beneficial for wider and/or deeper models (that is, typical setups where MP is used).

\begin{table}[ht]
\caption{  DCT-MP on a translation model with IWSLT'14 dataset: Train and Test Losses and BLEU scores for various levels of sparsity and different splits for MP.}
\label{table:DP_MP_NLP}
\begin{center}
\begin{small}
\begin{sc}
\begin{tabular}{lcccr}
\toprule
\makecell{Sparsity \\ Factor} & 
\makecell{MP \\ Workers} & 
\makecell{Train\\Loss } &
\makecell{Test\\Loss} &
\makecell{BLEU \\Score} 
\\
\midrule
Baseline  & -- & 3.150 & 3.883 &  35.17 \\
\cmidrule{1-5}
90\%      & 2 & 3.159 & 3.879 &  35.23\\
\bf 95\%      & 2 & 3.157 & \bf 3.882 &  35.18\\
\cmidrule{1-5}
90\%      & 3 & 3.151 & 3.881 &  35.22\\
\bf 95\%      & 3 & 3.148 & \bf 3.882 &  35.19\\
\cmidrule{1-5}
\bf 90\%      & 5 & 3.157 & \bf 3.882 &  35.20\\
95\%      & 5 & 3.188 & 3.890 &  35.15\\

\bottomrule
\end{tabular}
\end{sc}
\end{small}
\end{center}
\end{table}

In this subsection, we do not consider DCT-DP since similar schemes have been evaluated for NLP models in existing works such as \cite{DP_survey} and \cite{aji2017sparse}. 
In the next subsection, we evaluate DCT-MP and DCT-DP on large-scale recommendation models for end-to-end training times and overall model performance.

\subsection{Large-Scale Recommendation System}
\label{sec:exps_prod}

\begin{figure*}[t]
    \centering
    \begin{subfigure}[t]{0.25\textwidth}
        \centering
        \includegraphics[width=\textwidth]{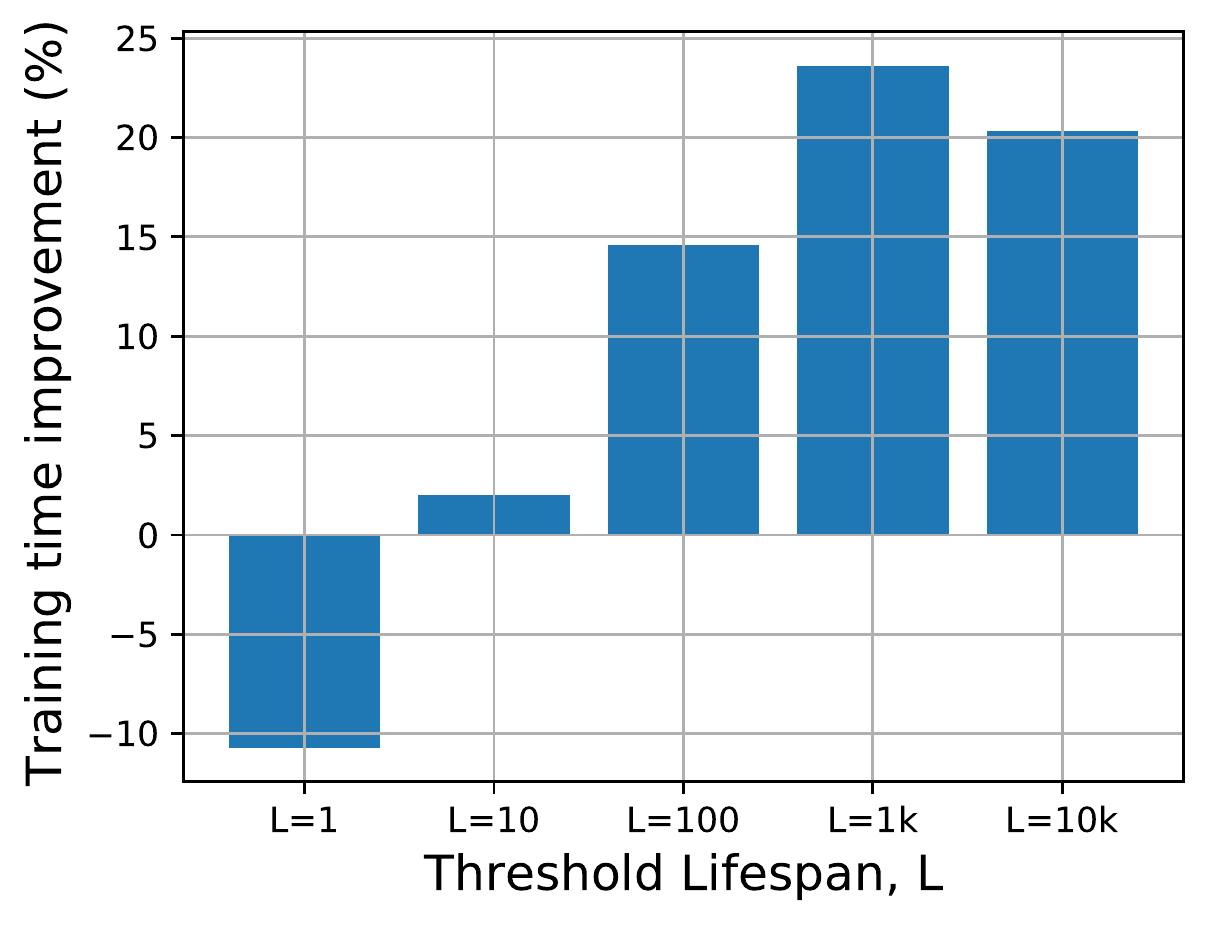}
        \caption{
        Training time improvements for different values of $L$. 
        }
    \label{fig:qps_l_perc}
    \end{subfigure}
    ~
    \begin{subfigure}[t]{0.25\textwidth}
        \centering
        \includegraphics[width=\textwidth]{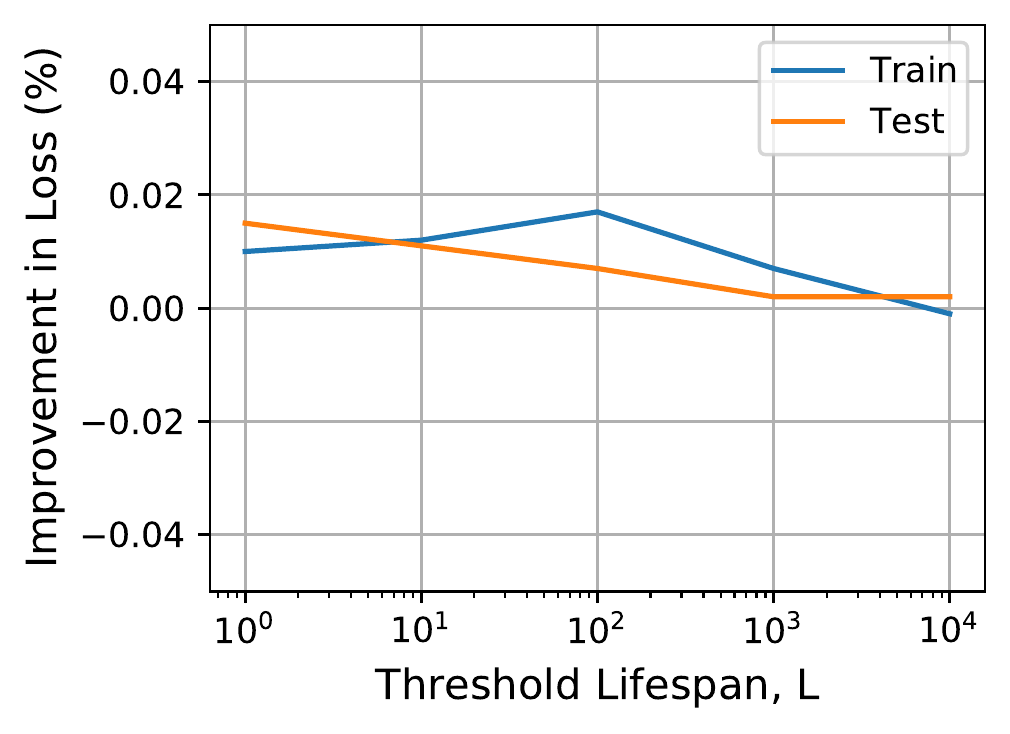}
        \caption{
        Train and test error improvements for different values of $L$.
        }
    \label{fig:train_test_L}
    \end{subfigure}
 ~   
    \begin{subfigure}[t]{0.25\textwidth}
        \centering
        \includegraphics[width=\textwidth]{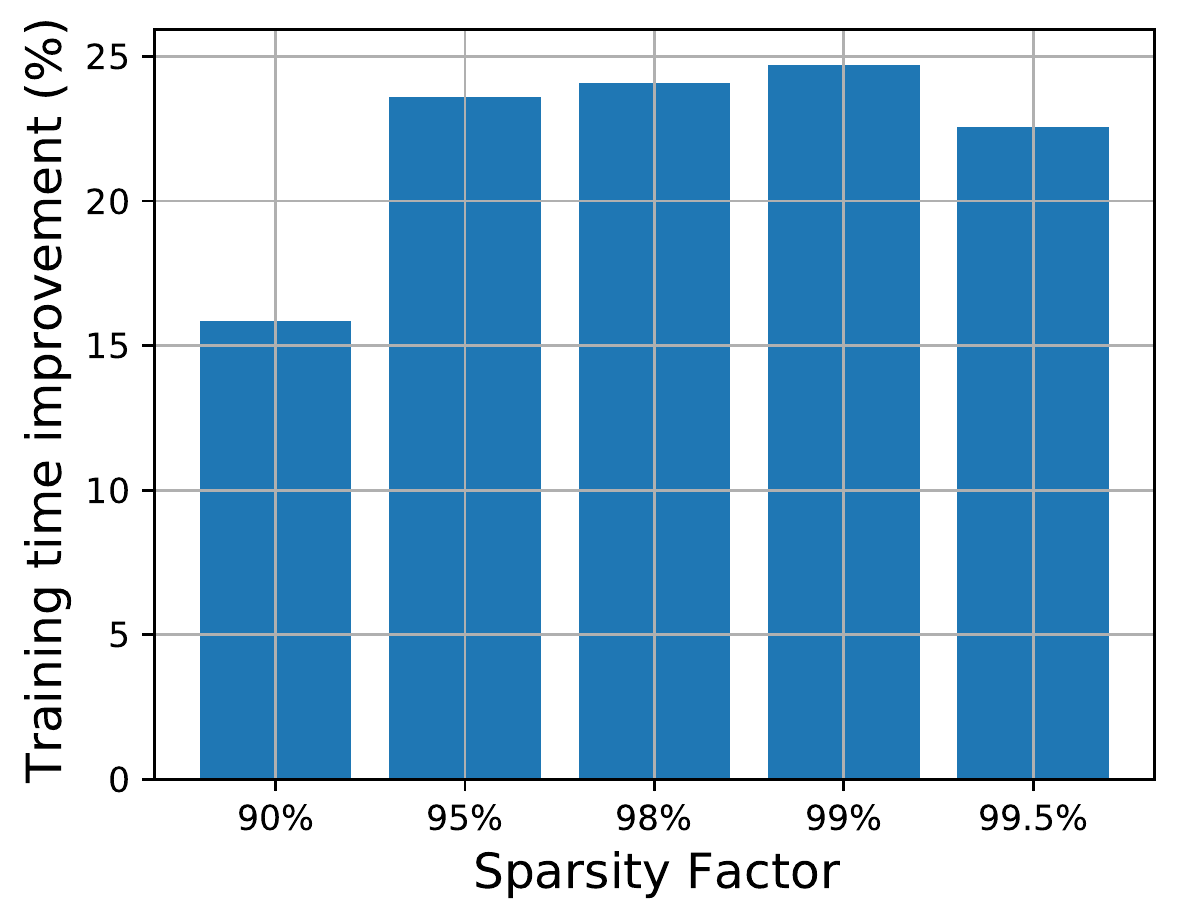}
        \caption{Training time improvements for various sparsity factors. 
        }
    \label{fig:qps_eta_perc}
    \end{subfigure}
    ~
    \begin{subfigure}[t]{0.25\textwidth}
        \centering
        \includegraphics[width=\textwidth]{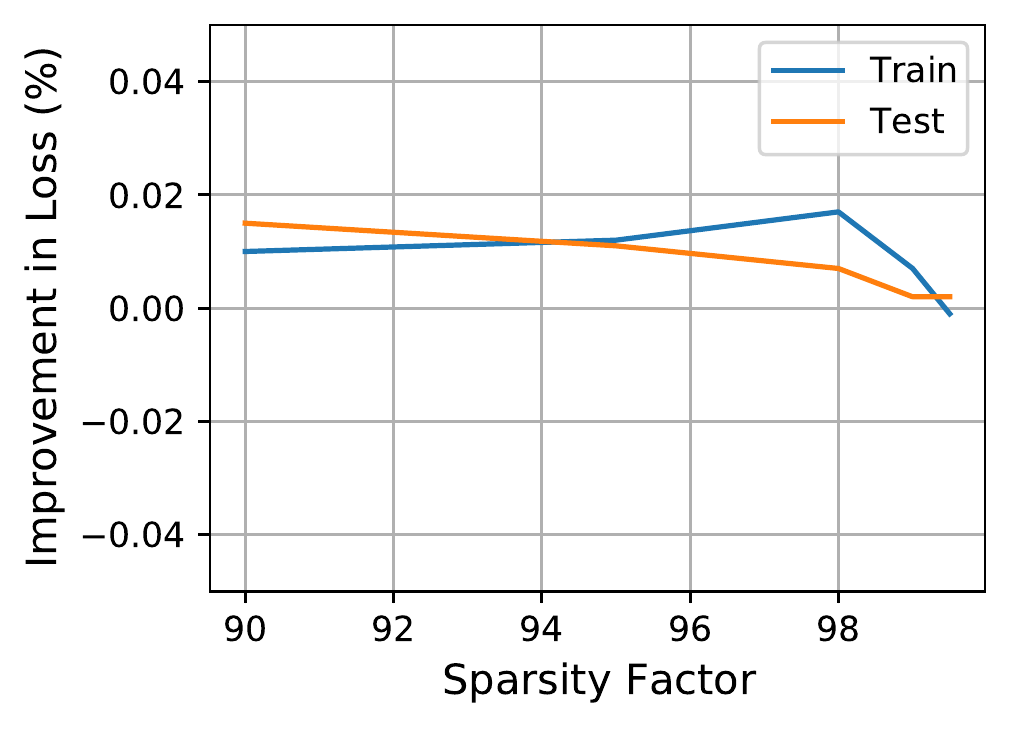}
        \caption{Train and test error for various sparsity factors.}
    \label{fig:train_test_eta}
    \end{subfigure}
    \caption{
    {DCT-DP on Large-Scale Recommendation Models. Figures (a) and (b) show the training time and loss improvements, respectively, over baseline for different values of the threshold life-span, $L$, for a sparsity level of $95\%$. Figures (c) and (d) show the same statistics for various levels of sparsity for $L=1000$.}
    }
\label{fig:DCT-DP-Prod}
\end{figure*} 

We present our results for a real-world large scale recommendation system that employs HP for parallelization on click-through rate prediction task. We employ DCT-MP and DCT-DP to reduce the network bandwidth usage in these systems.

{\bf Experimental Setup.}
We leverage a distributed data-parallel asynchronous training system with multiple trainers to train a recommendation model. Each trainer in the DP setup may consist of one or more workers that use MP (see Fig. \ref{fig:hybrid_illus} for an illustration). Typically, the model is split into 10 or more parts and fine-grained parallelism is employed for high throughput. Hence, the worker machines suffer from very high communication cost for both MP and DP. 
The batch sizes are usually in the range of 100-1000, but they are employed with hogwild threads (see \citet{hogwild}) to increase the throughput of the system, further exacerbating the communication cost problem. 
The recommendation model considered in this section takes multiple days to train with general-purpose CPU machines. All the workers and parameter servers run on Intel 18-core 2GHz processors with 12.5Gbit Ethernet. The hardware configurations are identical and consistent across all the experiments.
We train using 7B training examples and evaluate the model on 0.5B examples. 
For quantifying model performance, we report the cross-entropy loss from the classification task. We compare relative cross-entropy loss and end-to-end training times of the proposed techniques with respect to a baseline model without communication compression.

{\bf DCT-DP with Large-Scale Recommendation Model.}
Figure \ref{fig:DCT-DP-Prod} shows the results of applying DCT-DP on the large-scale recommendation model. 
In Figure \ref{fig:qps_l_perc}, we plot the improvements in end-to-end training times when DCT-MP is applied to compress the parameter gradients, $W_{grad}$, that are sent to the parameter server. Here, we keep the sparsity level constant at $95\%$ and vary the threshold life-span $L$ (the interval after which the top-K threshold is updated).  We note that compression with $L=1$ takes $11\%$ more time than the baseline with no compression. This is due to the cost of the copy-and-sort routine which computes the top-K threshold.\footnote{Note that $L=1$ represents the scheme proposed in popular works such as \cite{stich_sparsified_sgd} and \cite{alistarh2018_conv_sparse}. Thus, naively implementing existing schemes for top-K sparsification might not always yield expected gains in production. However, as we observe later, simply updating $L$ every thousand iterations can improve the training time by $25\%$ without any loss in performance.} Increasing $L$ to $1000$ trains the model $23\%$ faster and further increasing it to $10000$ does not provide any additional gain. Figure \ref{fig:train_test_L} illustrates that for different values of $L$, the train and test losses are within $0.01\%$ of the baseline~performance.

Fig. \ref{fig:qps_eta_perc} shows the improvement in training time for various levels of sparsity when the threshold life span is kept constant at $L=1000$. We observe the general trend that when the sparsity is increased, the training time improves. 
Overall, we are able to compress the gradients to sparsity factors of up to $99.5\%$ without any loss in train and test performance (as noted from Fig. \ref{fig:train_test_eta}). However, we do not see significant improvements in training time beyond the sparsity level of $95\%$, possibly because the message size is small enough to not hurt bandwidth usage, and the only cost remaining is the fixed latency cost associated with sending any message, irrespective of its size.
\begin{remark}
We observe that error feedback works very well in this asynchronous data-parallel training paradigm with a larger number of hogwild threads.
 Note that this should not be expected since existing works prove convergence guarantees only for the synchronous SGD settings.
An implementation detail that helped was sharing the error feedback buffer between the multiple threads. But this can lead to a fast growing magnitude of error in the buffer leading to stale updates. To avoid this, we drain the error feedback buffer stochastically every $1$ million iterations. 
\end{remark}


{\bf DCT-MP with Large-Scale Recommendation Model.}
We employ DCT-MP to compress the entities sent through the network during MP for communication efficiency. DCT-MP is applied across the 12 splits of the model after the ReLU layer. Our results are summarized in Table \ref{table:prod_mp}. 
We show improvement in training and test losses%
\footnote{Positive numbers imply better performance.} 
in columns 2 and 3, respectively, and the improvements in end-to-end training times in column 4 for various levels of sparsity. 
We observe that the training performance slightly degrades with DCT-MP on large-scale models.
However, the test performance improves up to sparsity levels of $90\%$, with a $14\%$ improvement in end-to-end training time. Increasing the sparsity level to $95\%$ degrades the test performance by $0.121\%$. Note that we can further improve the performance of DCT-MP by identifying the layers whose activations are sensitive to sparsification and avoiding compressing them during DCT-MP (or changing the location of the split). However, such selectivity in choosing layers for DCT-MP is beyond the scope of this paper.

\begin{table}[t]
\caption{  DCT-MP on a large-scale recommender model}
\label{table:prod_mp}
\begin{center}
\begin{small}
\begin{sc}
\begin{tabular}{lccr}
\toprule
\makecell{Sparsity \\ Factor} & 
\multicolumn{2}{c}{
\makecell{Loss Improvement (\%)\\Train \phantom{iiiiiiii} Test}} &
\makecell{Time \\Gain (\%)} 
\\
\midrule
Baseline  & 0.000\%  & 0.000\% &   0.000\% \\
75\%      & -0.006\% & 0.023\% &   7.04\%\\
\bf 90\%      & -0.021\% & \bf 0.016\%  & \bf 13.95\%\\
95\%      & -0.070\% & -0.121\%  &  14.43\%\\
\bottomrule
\end{tabular}
\end{sc}
\end{small}
\end{center}
\end{table}

{\bf Communication-Efficient Hybrid training.}
Next, we apply both DCT-DP and DCT-MP for communication reduction during hybrid training of a large-scale recommendation model. Inspired by our previous results, we chose the sparsity levels as $90\%$ and $99\%$ for DCT-MP and DCT-DP (with $L=1000$), respectively. We observe a $37.1\%$ reduction in end-to-end training time, with train and test loss within $0.01\%$ of the baseline model that does no compression. 

Further, before applying DCT, we observed that the network utilization was high (94.2\%) and the CPU utilization was low (48.7\%), implying that communication is a bottleneck. However, after applying DCT, CPU utilization increased to 91.1\% and network utilization decreased to 49.3\%, implying that DCT shifted the bottleneck from communication to computation in production models. 
\section{Conclusions}

Inspired by the fact that communication is increasingly becoming the bottleneck for large-scale training, we proposed two practical algorithms, DCT-DP and DCT-MP, to reduce the communication bottleneck during data and model parallelism, respectively, for fast training of DNN models. DCT-DP and DCT-MP improve end-to-end training time by sparsifying the matrices to be sent across the wire by appropriately selecting a sparsification threshold. We empirically evaluated the proposed algorithms on publicly-available as well as industry-scale models and datasets. We show a reduction in communication for MP and DP by up to $20\times$ and $100\times$, respectively, without any loss in performance.
Further, the end-to-end training time reduces by $37\%$ in production models. Further, our algorithms reduce the network bandwidth utilization by half and almost double the CPU utilization, shifting the training bottleneck from communication to computation.  



\bibliographystyle{ACM-Reference-Format}
\bibliography{bibli}

\clearpage

\appendix

\section{Schemes That Do Not Work}
\label{app:sketch}

We saw in Sec. \ref{sec:exps} that activations during the forward pass and gradients during the backward pass can be compressed by large factors (up to $20\times$) using DCT-MP. This is due to selecting and training only the most relevant neurons corresponding to a given training sample. In this section, we present some negative results with other methods to compress the activation and gradient matrices during the forward and backward passes, respectively. 

{\bf Gaussian Sketching for Activation Compression.} 
Here, we use a Gaussian sketching scheme to compress the activations going forward. In Randomized Numerical Linear Algebra (RandNLA), the idea of sketching is to represent a large matrix by a smaller proxy that can be further used for matrix operations such as matrix multiplication, least squares regression, and low-rank approximation \cite{woodruff_now}. The sketched version of a matrix $A$ is given by $A\times S$, where $S$ is a random sketching matrix (e.g., all entries of $S$ are sampled i.i.d. from an appropriately scaled Gaussian distribution).

In Table \ref{table:MP_gaussian_sketching}, we compress the activations during the forward pass using Gaussian sketching. 
Unlike the DCT-MP algorithm, we do not compress the gradients during the backward pass. 
The aim is to identify if a low-rank structure exists in the activation matrix that can be used to compress the activation matrix in general.


\begin{table}[h]
\caption{Compressing the activation matrix during MP using Gaussian sketching does not yield good results.}
\label{table:MP_gaussian_sketching}
\begin{center}
\begin{small}
\begin{sc}
\begin{tabular}{lcccr}
\toprule
\makecell{Compression \\ Factor} & 
\multicolumn{2}{c}{
\makecell{Train\\Loss \phantom{iii} Acc (\%)}} &
\multicolumn{2}{c}{
\makecell{Test\\Loss \phantom{iii} Acc (\%)}} \\
\midrule
Baseline  & 0.4477 & 79.23 & 0.4538 &  78.78\\
50\%      & 0.4569 & 78.72 & 0.4618 &  78.37\\
75\%      & 0.4610 & 78.53 & 0.4656 &  78.12\\
90\%      & 0.4685 & 77.95 & 0.4721 &  77.78\\
\bottomrule
\end{tabular}
\end{sc}
\end{small}
\end{center}
\vskip -0.1in
\end{table}

As seen in Table \ref{table:MP_gaussian_sketching}, sketching techniques directly borrowed from RandNLA do not perform as well. 
This is likely because such schemes were designed to cater to operations such as low-rank approximation, where the matrices to be compressed are generally well-approximated by low-rank matrices. 
For instance, Gaussian sketching has seen success in approximate least squares regression and low-rank matrix approximation \cite{woodruff_now}. 
This suggests that the activation matrix for DNNs, in general, does not reside in a subspace that is sufficiently low-rank to be meaningfully used for compression.


{\bf Top-K Thresholding for Gradient Compression.} 
We saw in Sec. \ref{sec:exps} that the parameter gradients (illustrated as $W_{grad}$ in Fig. \ref{fig:hybrid_illus}) can be compressed to high factors with any loss in accuracy when used with appropriate error compensation. However, the same is not true for the gradients with respect to hidden neurons (illustrated as $X_{grad}$ in Fig. \ref{fig:hybrid_illus}) that are sent across the network during the backward pass in MP. This can be seen from our results in Table \ref{table:MP_grad_topK}, where we apply gradient compression using top-K thresholding with error feedback. Further, we observed that training without error feedback can cause~divergence.

\begin{table}[h]
\caption{Compressing the gradient matrix during backward pass in MP using top-K sparsification does not yield good results.}
\label{table:MP_grad_topK}
\begin{center}
\begin{small}
\begin{sc}
\begin{tabular}{lcccr}
\toprule
\makecell{Compression \\ Factor} & 
\multicolumn{2}{c}{
\makecell{Train\\Loss \phantom{iii} Acc (\%)}} &
\multicolumn{2}{c}{
\makecell{Test\\Loss \phantom{iii} Acc (\%)}} \\
\midrule
Baseline  & 0.4477 & 79.23 & 0.4538 &  78.78\\
50\%      & 0.4495 & 79.07 & 0.4561 &  78.62\\
75\%      & 0.4516 & 78.95 & 0.4588 &  78.48\\
90\%      & 0.4701 & 77.76 & 0.4789 &  77.13\\
\bottomrule
\end{tabular}
\end{sc}
\end{small}
\end{center}
\vskip -0.1in
\end{table}

Our hypothesis on why compressing the gradients of the hidden neurons by top-K thresholding does not yield good results is due to the propagation of error to the initial layers. 
Consider the following example to illustrate this.
Consider the following deep network, where we have several vector-valued functions $A(\cdot), B(\cdot), C(\cdot), \cdots, L(\cdot)$  composed in a chain, that is $A \rightarrow B \rightarrow C \rightarrow \cdots \rightarrow K \rightarrow L$. 
Algebraically, the loss looks like $L(A) = L(K(\cdots C(B(A))\cdots))$. 
Then, the gradient of the loss with respect to $A$ is given by the multiplication of the Jacobians, that is, $J_L(A) = J_L(K) \times \cdots \times J_C(B) \times J_B(A)$. 
(Here, $J_L(A)$ denotes the gradient of $L$ with respect to $A$.) 
If we change any of the Jacobians in between (that is, compress the gradient $X_{grad}$ with respect to hidden neurons), then the error is propagated all the way to the initial layers of the network. 
Even adding error feedback to the compression process does not recover the lost accuracy.

\section{Analysis of DCT-MP (Algorithm \ref{alg:wta_mp})}
\label{app:analysis_dct_mp}

Unlike compression of gradients, where the error introduced can be effectively corrected, compressing activations introduces errors that are propagated to the downstream of the network. 
If the thresholds $\{\tau_i\}_{i=1}^B$ are fixed for all training iterations, then Algorithm \ref{alg:wta_mp} is simply performing SGD of a changed network, one in which the additional layer $\tilde{X}_{act,i} = \I(|X_{act,i}| \geq \tau_i)$
inserted right after $\{X_{act,i}\}_{i=1}^B$ are obtained. 
Extra analysis is needed when $\{\tau_i\}$ are dynamic.
Consider a particular SGD iteration $k$, and further annotate the thresholds as $\{\tau_i^{(k)}\}$ to indicate their dependence on the stochastic mini-batch being chosen at iteration $k$. 
Let $\bar{\tau}^{(k)} = \mathbb E_i(\tau_i^{(k)})$ be the average of these thresholds over the entire dataset. 
Denote by $\bar{L}$ the loss function of the network with this batch-independent threshold $\tilde{X}_{act,i} = \I(|X_{act,i}| \geq \bar\tau^{(k)})$ inserted, and let $L_k$ be the loss function of this dynamically thresholded network with batch $k$. 
Standard SGD assumptions say that for a randomly chosen batch $k$, $\mathbb E_i(\partial L_i/\partial \bm{\theta}) = \partial L/\partial \bm{\theta}$, with $\bm{\theta}$ being the network parameters. 
However, since each $L_i$ is related to a different threshold, it is unclear what the $L$ should be on the right hand side of this expression. 
In the following theorem, we show that $\mathbb E_i(\partial L_i / \partial \bm{\theta}) = \partial \bar{L}/\partial \bm{\theta}$, where $L_i$ is batch-$i$ loss function of the \emph{dynamically} thresholded network in Algorithm \ref{alg:wta_mp}.

\begin{theorem}
\label{thm:unbias}
Consider a 2-worker MP network where the activations from Worker 1 to Worker 2 are thresholded as in Algorithm \ref{alg:wta_mp}. 
Let $L_i$ be the associated loss function for data point $i$, where $i=1,2,\ldots,N$, and $N$ be the total number of training samples. 
Let $\bar{\tau} = \E_i(\tau_i)$, the entire data set at a particular training iteration $j$ (the subscript $j$ in $\tau_i$ and $\bar{\tau}$ is omitted for simplicity). 
Let $\bar{L}_i$ be the loss function corresponding to the threshold $\bar{\tau}$. 
If
\begin{equation}\label{eq:assumption}
\E_i[X_{act,i} \odot \I(|X_{act,i}| \ge \tau_i) ]
= \E_i[X_{act,i} \odot \I(|X_{act,i}| \ge \bar{\tau}) ],    
\end{equation}
then, up to first order 
\[
\E_i\left(\frac{\partial L_i}{\partial \bm{\theta}}\right) =
\E_i\left(\frac{\partial \bar{L}_i}{\partial \bm{\theta}}\right) = 
\frac{\partial \bar{L}}{\partial \bm{\theta}}
\]
where $\bm{\theta}$ is the parameter of the network. 
\end{theorem}

\begin{remark}
The theorem statement implies that the training step on the dynamically thresholded network is equivalent to that of training the network with just one threshold $\bar{\tau}$. This happens long as $\bar \tau$ and $\tau_i$ are sufficiently close and the mean of activations around $\tau_i$ is the same as their mean around $\bar\tau$ (formalized by assumption (\ref{eq:assumption})).
\end{remark}

\begin{proof}
To establish the proof, let the dynamically thresholded network be represented as follows, starting with a random data point $(X_i^{(0)},y_i)$ (where $X_i^{(0)}$ is the input and $y_i$ the corresponding label):
\begin{eqnarray*}
    X_i^{(k)} & = & \cN^{(k)}(\bm{\theta}^{(k)},X_i^{(k-1)}), \quad k = 1,2, \ldots, m \\
    X_i^{(m+1)} & = & X_i^{(m)} \cdot \I(|X_i^{(m)}| \ge \tau_i) \\
    X_i^{(k)} & = & \cN^{(k)}(\bm{\theta}^{(k)},X_i^{(k-1)}), \quad k = m+2, m+3, \ldots, K\\
    L_i &= & \ell(X_i^{(K)},y_i).
\end{eqnarray*}
Here, the MP split was inserted after the $k$-th layer, and the activation function for the $k$-th layer with parameters $\bm{\theta^{k}}$ is represented as $\cN^{(k)}(\bm{\theta^{k}}, \cdot)$. For an activation layer $\cN^{(k)}$, $\bm{\theta}^{(k)}$ is simply an empty set. For the network with a static threshold~$\bar{\tau}$:
\begin{eqnarray*}
    \bar{X}_i^{(m+1)} & = & X_i^{(m)} \cdot \I(|X_i^{(m)}| \ge \tau) \\
    \bar{X}_i^{(k)} & = & \cN^{(k)}(\bm{\theta}^{(k)},\bar{X}_i^{(k-1)}), \quad k = m+2, m+3, \ldots, K\\
    \bar{L}_i &= & \ell(\bar{X}_i^{(K)},y_i).
\end{eqnarray*}

By assumption (\ref{eq:assumption}), we have 
$\E_i(X_i^{(m+1)}) = \E_i(\bar{X}_i^{(m+1)})$. 
Therefore, by Taylor's expansion, we have
\[
\E_i(X_i^{(k)}) = \E_i(\bar{X}_i^{(k)}), \quad k = m+2, m+3, \ldots, K,
\]
up to first order, by expanding each $\cN^{(k)}(\bm{\theta}^{(k)},\bar{X}_i)$ to first order. For example, for a linear layer, we have
\begin{eqnarray*}
\E_i(X_i^{(k)}) & = & \E_i( \cN(\bm{\theta}^{(k)},X_i^{(k)}) )\\
    & = & \E_i( W^{(k)} X_i^{(k)} + b^{(k)} ) \\
    & = & W^{(k)} \E_i(X_i^{(k)}) + b^{(k)} \\
 & = & \bar{X}_i^{k+1} .
\end{eqnarray*}
Similarly, for a general activation function $\sigma(\cdot)$, we have
\begin{align*}
& \E_i(\sigma(X_i^{(k)})) = \E_i(\sigma(\bar{X}_i^{(k)} + (X_i^{(k)}-\bar{X}_i^{(k)}) )) \\
& = \E_i(\sigma(\bar{X}_i^{(k)}) + \sigma'(\bar{X}_i^{(k)})\cdot (X_i^{(k)}-\bar{X}_i^{(k)})) + \hbox{second order terms} \\
& = \E_i(\sigma(\bar{X}_i^{(k)})) + \sigma'(\bar{X}_i^{(k)}) \E_i(X_i^{(k)}-\bar{X}_i^{(k)}) \\
& = \E_i(\sigma(\bar{X}_i^{(k)})) = \E_i(\bar{X}_i^{(k+1)}),
\end{align*}
where we have ignored the second-order terms beyond the second step.
Let $\Delta_i^{(k)} = X_i^{(k)} - \bar{X}_i^{(k)}$, $E_i(\Delta_i^{(k)}) = \mathbf{0}$. Consider the gradient $\frac{\partial L_i}{\partial \bm{\theta}}$ as a function of $\Delta_i^{(k)}$:
\begin{equation*}
\frac{\partial L_i}{\partial \bm{\theta}}(\Delta_i^{m+1},\Delta_i^{(m+2)},\ldots,\Delta_i^{(K)})
= \frac{\partial \bar{L}_i}{\partial \bm{\theta}} + 
\sum_{k=m+1}^K \left(\frac{\partial^2 \bar{L}_i}{\partial \Delta^{(m)} \partial \bm{\theta}} \right) \cdot \Delta_i^{(k)}.
\end{equation*}
Thus
\[
\E_i\left( \frac{\partial L_i}{\partial \bm{\theta}} \right) =
\E_i\left( \frac{\partial \bar{L}_i}{\partial \bm{\theta}} \right),
\]
which proves the desired result.
\end{proof}

\end{document}